\documentclass[letterpaper, 10 pt, conference]{ieeeconf}  
                                                          \usepackage{subfiles}
\usepackage{graphicx}
\graphicspath{ {images/} }
\usepackage{caption}
\usepackage{subcaption}
\usepackage{lipsum}
\usepackage{float}
\IEEEoverridecommandlockouts                              
\usepackage{graphics} 
\usepackage{epsfig} 
\usepackage{mathptmx} 
\usepackage{times} 
\usepackage{amsmath} 
\usepackage{amssymb}  
\usepackage{booktabs}
\title{\LARGE \bf
Bitwise Operations of Cellular Automaton on Gray-scale Images
}

\author{Karttikeya Mangalam$^{1}$ and Prof. K S Venkatesh$^{2}$
\thanks{*This work was not supported by any organization}
\thanks{$^{1}$Karttikeya Mangalam is pursuing B.Tech. at Department of Electrical Engineering, Indian Institute of Technology, Kanpur, India
        {\tt\small mangalam at iitk.ac.in}}%
\thanks{$^{2}$Prof. K S Venkatesh is with the Department of Electrical Engineering, Indian Institute of Technology, Kanpur, India.
        {\tt\small venkats at iitk.ac.in}}%
}

\begin{document}

\maketitle
\thispagestyle{empty}
\pagestyle{empty}

\begin{abstract}
Cellular Automata (CA) theory is a discrete model that represents the state of each of its cells from a finite set of possible values which evolve in time according to a pre-defined set of transition rules.
CA have been applied to a number of image processing tasks such as Convex Hull Detection, Image Denoising etc. but mostly under the limitation of restricting the input to binary images. In general, a gray-scale image may be converted to a number of different binary images which are finally recombined after CA operations on each of them individually.  
We have developed a multinomial regression based  weighed summation method to recombine binary images for better performance of CA based Image Processing algorithms.
The recombination algorithm is tested for
the specific case of denoising Salt and Pepper Noise to test against standard
benchmark algorithms such as the Median Filter for various images and noise levels. The results indicate several interesting invariances in the application of the CA, such as the particular noise realization and the choice of sub-sampling of pixels to determine recombination weights. Additionally, it appears that simpler algorithms for weight optimization which seek local minima work as effectively as those that seek global minima such as Simulated Annealing. 

\end{abstract}

\section{INTRODUCTION}
The idea of the Cellular Automaton in its present form was conceptualized by J. Von Numann and Stan Ulam \cite{foundations1}. A rich theory for CA was worked out by Steven Wolfram \cite{foundations2}. For completing computation heavy tasks in real time, parallel computing is one of the brightest hopes and CA models lend very easily to parallel implementation due to the inherent independence in their model structure \cite{deepusurvey}. Image processing requires heavy computation in realtime and thus CA is a very intuitive and natural candidate for the task. \\
Digital image processing is an integral part of modern day signal processing, with applications ranging from Magnetic Resonance Imaging to Satellite Television. One of the major problems faced is the corruption of images by various noise sources. Out of the many possible noises, impulse noise, such as Salt and Pepper, is a common challenge in image processing: it is routinely caused by a number of possible aberrations such as malfunctioning of sensors, noisy transmission over a channel leading to loss of some bits, or corruption in specific memory locations \cite{selva}.
\\
A standard approach to reduce salt and pepper noise is the use of median filters \cite{median1}\cite{median2}. Popvici et al. designed a local transition function such that the current cell's state is a function of the previous states of the neighboring cells \cite{Pops}. However, these methods are not optimal since, these are agnostic to the presence of local features like edges and hence output blurry images especially for high noise levels. \cite{caforip1} provides a CA algorithm to deal with higher noise levels. Rosin et al. propose a greedy approach to rule selection that includes denoising salt and pepper noise \cite{ble2} using CA. Although each cell only sees and interacts with a few other cells according to some simple rules, the combination of these local interactions can cause complex non-linear global behavior as seen in the famous example of Conway's Game of Life \cite{conway}. \\ 
Due to these promising advancements in the use of CA for denoising Salt and Pepper noise, we have chosen it as a model for demonstrating our general algorithm for extending CA methods to gray scale images. Moreover, it allows for comparison against standard benchmark algorithms such as the median filter. \\
A major hurdle for wider application of CA theory is the limitation to a very small number of allowable states for the image. Generally, the number of allowable states is limited to just two or three because of the exponential dependence of the size of the search space of allowable rule-sets on the number of possible states for each cell. This makes the problem of finding suitable rule-sets for any application very computationally intensive, especially when rather larger numbers of possible states need to be condsidered. In our work, we propose to overcome this hurdle by actually constraining the search space for rule sets to only two possible states (Binary images) for each pixel for each CA in use, but still providing for  application of CA to Grayscale images that actually have 256 possible states for each pixel.\\
Section II describes the CA framework for application of the Recombination Algorithm. The first sub-section introduces the general CA setting and the standard definitions and terms. The second sub-section narrows the setting to that of binary images and finally, the last sub-section describes the step-wise algorithm for extension of Binary methods to gray-scale images. The second section introduces our proposed Recombination Algorithm which is followed by the simulation results and discussion. We end with acknowledgments and referencesmakemytrip
.
\\

%
%


\section{CA MODEL FOR DIGITAL IMAGES}
 
 A Cellular Automaton model for operations on images is essentially defined by a two dimensional grid like structure with each cell belonging to a particular state that is chosen from a set of finitely many choices at each time step. Moreover, this state changes in time according to a well defined set of rules governing the 'evolution' of CA which might take into account a history of all the previous states for both the cell and its neighbors. This 'rulebook' is generally taken to be time-invariant and independent of the cell in question \cite{foundations2}.

\subsection{Model structure and parameters}

A two dimensional deterministic CA can be understood and represented as an ordered three tuple $\langle S,B^{1}_r,\Delta \rangle$ where $S$ is the State set. A standard way to define neighborhood $B^{1}_r \in \mathbf{Z^2} $ is to use a ball defined by a metric $B^{1}_r(\bf{x},\bf{x_c})$ \cite{Pops}. Different metrics lead to a variety of neighborhoods such as Von Neumann Neighborhood:
\begin{equation}
B^1_r({\bf x_c}) := \{ {\bf x}: \vert {\bf x} - {\bf x_c} \vert_1 \le r\}; \ {\bf x}=  \langle x,y \rangle
\end{equation}
Similarly, Moore Neighborhood is defined by the norm \cite{mathe1}

\begin{equation}	
B^{\infty}_r({\bf x_c}) := \{ {\bf x}: \vert {\bf x} - {\bf x_c} \vert_\infty \le r\}; \ {\bf x}=  \langle x,y \rangle
\end{equation}
 $\Delta$ is a function that has the argument as current state of the given cell and its neighborhood and returns the next state of the given cell, according to a deterministic rule. The set of states, the neighborhood patterns and the rule set ($\Delta$) along with the initial states completely define a certain CA.

\begin{figure}
\begin{center}
\begin{tabular}{ll}
\includegraphics[width = 0.1\textwidth]{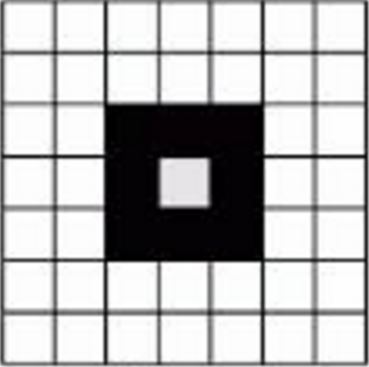}
&
\includegraphics[width = 0.1\textwidth]{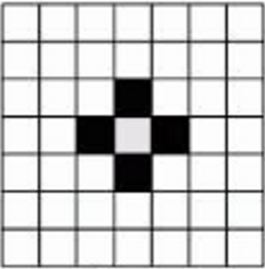}
\end{tabular}
\end{center}
\caption{Moore (Left) and Von Neumann neighborhood (Right)}
\end{figure}

\subsection{Application of CA to Binary Images}
A binary image in CA framework is modeled as a 2-dimensional binary state CA with a Moore neighborhood of $r$ = 1 or 2 in most cases. In general, the inverse problem of finding the rule set for use of CA to perform a given image processing task is highly computationally intensive. For a CA with $S$ states and a neighbourhood consisting of $N$cells $\Delta$ has a total of $S^{S^N}$ possible choices. Even with a case as small as that of Moore neighborhood with $R$ = 1 ($N$ = 8) with binary states, there are more than $10^{77}$ possible rulesets. This exponential dependence makes it pragmatically computationally impossible to evaluate every possible ruleset combination for any given task and choose the one with the best performance. \\
Many different tricks can be employed for working around this problem. The most employed of them is the restriction of input to binary images ($S$=2). The reason for this is to limit the number of possible rulesets to as small as possible. Still, since the possible number is quite large, some researchers have chosen these rulesets heuristically by hand \cite{frmble2}. On the other hand, Rosin et al. have applied a general framework known as Sequential Forward Floating Search (SFFS) for greedily identifying the rulesets that perform optimally at each step \cite{ble2}. 
\subsection{Extension to Grayscale Images}
Restricting the input to binary images poses the problem of application of CA to grayscale images. One of the common strategies applied is to convert the original grayscale $m \times n$ image to a number of different binary $m \times n$ images. Then, we apply CA individually to each of these binary images and finally recombine them to obtain the final grayscale image.  Mathematically, these three steps can be described as the following operations: 
\vspace{2 mm}
\subsubsection*{Step 1: Thresholding the Grayscale Image}
In the first step, for conversion to binary images, a number of strategies are used. A binary threshold can be introduced at each value of set $K=\{k_i\}$ of chosen intensity values, which could even be all (256 in total) of the available grey levels. We define an auxiliary transformation $T_k$ as follows: 
\begin{equation*} 
\begin{aligned}
\label{eq1}
T_k : &\{\, X : \, X(i,j)\in (0,L)\, \} \mapsto \{\, I : I(i,j) \in \{0,1\}\, \}  ; k \in K
\\
\vspace{1mm}
& A_X(i,j,k) := T_k(X )(i,j)
\end{aligned}
\end{equation*}
Thus, $A_X$ is a $k$-indexed stack of $\vert K \vert$ binary images.  For example, one interpretation of the above would be
\begin{equation}
I_k(i,j) = :
 \begin{cases} 
      1 &  X(i,j) \geq k\\
      0 &  Otherwise
   \end{cases}
; k \in \{0, 1,2,\cdots, 255\}
\end{equation}
This would generate 256 binary images from each grayscale image. \\   Another  possible implementation is to threshold not at all the possible gray values, but only at the positive powers of 2 until 255. So, in the above notation then, 
 $k \in \{1,2,4,8,16,32,64,128\}$ , which are usually termed as the eight bitplane images. This has the advantage of reducing the number of binary images from 256 to only 8 possible images with a corresponding reduction in computation time. However, the loss of original intensity information here is too large and thus the resulting image is significantly blurred on recombination.\\ A third  possible method for decomposition is to divide the input image into a number of different non-overlapping patches and then compute OTSU thresholds for each of these patches separately \cite{otsu} and now use these computed thresholds as values of $k$ over the entire image. Though these computed thresholds are specific to the image, hardly any performance boost is achieved by cutting down on the number of thresholds using this method, because depending upon the number of non-overlapping regions selected, these thresholds can themselves become too many in number and present a more or less uniform distribution over the interval [0,255]. Thus, a good option for a thresholding step is to either threshold at every possible value  $k \in \{0,1,2,\cdots, 255\}$ or to uniformly sample half or one third of this complete set, for example $k \in \{1,4,7 \cdots, 253\}$ etc.
 \vspace{2 mm}
\subsubsection*{Step 2: Application of CA rules}
In this step, each binary image resulting from Step 1 is applied to a CA to obtain an equal number of output binary images. 
\begin{center}
$CA : \{ \, A_X \mapsto  A_X' \, \}$ 
\end{center} $A_X'$ is also a stack of $\vert K \vert $ binary images. \\ 
The selection of the rules is an entire problem in itself. \cite{pattern} works on using CA for pattern classification. \cite{ble1} and \cite{ble2} propose methods for generating rule sets for thinning, denoising, convex hull detection etc. \cite{selva} proposes a majority rule for denoising SPN noise. \cite{rotation} develops algorithm for arbitrary rotation of images using two dimensional CA. At Step 2, any of these rules maybe applied individually over each component binary image $I_k$ generated after Step 1. The selection of the particular CA rule depends upon the application at hand and the method being employed. Once selected, the rules are applied on individual pixels of the binary image $I_k$ and result is stored in the output images at the identical position.
For our demonstration in this paper, we have chosen salt and pepper denoising.
\vspace{-2mm}
\subsection*{Step 3: Recombination to Form the Output Grayscale Image}\vspace{-8mm}
\begin{equation} 
T'_k : \{\, A_X' \vert \, A(i,j,k)\in {0,1} \, \} \to \{\, X' \, |\, I(i,j) \in \{0,1\}\, \}
\end{equation}
This is where our work makes the most contribution. The recombination algorithm in general, depends upon the method employed for thresholding in Step 1, but our proposed method takes into account this complication and adjusts itself so as to deliver the optimal output. The trade off is a corresponding increase in computation time but that can be circumvented by pre-processing the image.

\section{Recombination Algorithm}
We propose a regression based weighed combination of binary images in Step 3 of the process. Suppose the chosen set of thresholds in Step 1 is $K = \{k_1,k_2, \ldots ,k_t\}$, then for $w_k > 0 $ we propose the recombination:
\begin{equation}
\sum\limits_{k=1}^{t}w_kA'_{X}(k) = X'
\end{equation}
where $A'_{X}$ is the binary image stack generated from applying CA operations to the $k$th binary image. $X'$ is our approximation of the desired output image. To find these weights $w_k$, we try to iteratively minimize the Mean Square Error between the uncorrupted pixels and the regenerated image $X'$. This approach assumes we have the knowledge of pixels that are uncorrupted by SPN noise which is easily gained by excluding the pixels having either very high (255) or very low (0) intensity values. Thus, the weight vector $W$ = $[w_1$,$w_2$, $\ldots$ ,$w_t]$ is obtained as:
\begin{equation}
W = \min\limits_{w_k \geq 0} \mathop{\sum\sum}_{(i,j) \in A*} \vert X'(i,j) - X(i,j) \vert^2  
\end{equation}
where $A$ is the set of all uncorrupted $(i,j)$. \\
A possible economization  is to uniformly sample a pre-decided percentage of pixels ($\eta$\%) from $A*$ and use that for calculating $W$ instead of complete $A$. So, 
\begin{equation}
W = \min\limits_{w_k \geq 0} \mathop{\sum\sum}_{(i,j) \in A_\eta} \vert X'(i,j) - X(i,j) \vert^2  
\end{equation}
The vector $W$ has to be appropriately randomly initialized for fast convergence. A good initialization is to use a set of uniform weights $w_k = \mu = 255/t ; k = {1,2, \ldots, |K|}$ because, this would have been the best estimate for weights had the CA operations not taken place. So, we may choose $w_k \sim U[(1-\epsilon)\mu , (1+\epsilon)\mu]$ for a small $\epsilon > 0$ such as 0.05. In practice, a value of $\eta$ = 0.1 was observed to be exactly as good as using the complete $A*$ for images down to 64 $\times$ 64 pixels in size. 
\section{Simulations}
\begin{figure}[t!]
\includegraphics[width = 0.5\textwidth]{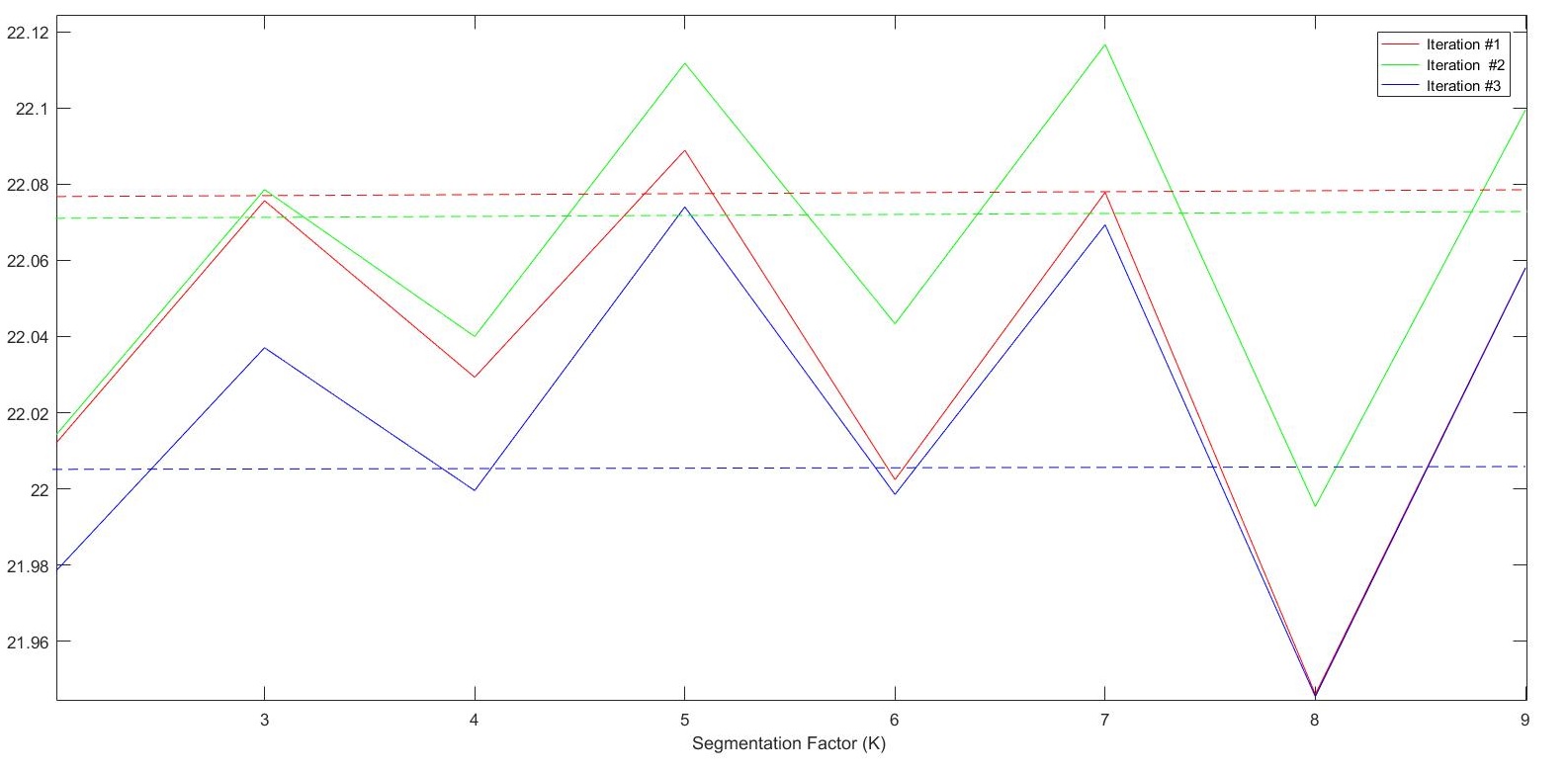}
\caption{Plot for Peak Signal to Noise Ratio (PSNR) Vs. Segmentation Factor for image for 3 iterations. Each Sub-sampling is performed on a separate realization of SPN Noise at noise level = 0.1. The dashed lines represent the PSNR value when all pixels are used for regression. A segmentation factor (k) means $\eta$ = $\frac{100}{k^2}$. 512x512 Lena Image is used.}
\end{figure}

\begin{table}[t!]
\centering
\begin{tabular}{@{}cccccc@{}}
\toprule
                                              & p = 0.06                   & p= 0.08                    & p = 0.1                    & p = 0.12                   & p = 0.14                   \\ \midrule
\multicolumn{1}{|c|}{Median Filtering}        & \multicolumn{1}{c|}{17.2}  & \multicolumn{1}{c|}{16.33} & \multicolumn{1}{c|}{15.42} & \multicolumn{1}{c|}{14.69} & \multicolumn{1}{c|}{13.81} \\ \midrule
\multicolumn{1}{|c|}{Using W = {[}1,1.. 1{]}} & \multicolumn{1}{c|}{18.94} & \multicolumn{1}{c|}{18.04} & \multicolumn{1}{c|}{17.34} & \multicolumn{1}{c|}{16.69} & \multicolumn{1}{c|}{15.8}  \\ \midrule
\multicolumn{1}{|c|}{$\eta$ = 100\%}             & \multicolumn{1}{c|}{20.1}  & \multicolumn{1}{c|}{19.98} & \multicolumn{1}{c|}{19.63} & \multicolumn{1}{c|}{19.34} & \multicolumn{1}{c|}{18.65} \\ \midrule
$\eta$ = 10\%                                    & 20.1                       & 19.98                      & 19.63                      & 19.33                      & 18.64                      \\ \bottomrule
\end{tabular}
\caption{PSNR Values achieved with various methods at different noise levels for SPN noise. 128x128 Lena Image is used for all comparisons.}
\label{my-label}
\end{table}

\begin{table}[t!]
\centering
\begin{tabular}{@{}lllllll@{}}
\toprule
              & MEAN   & MEDIAN & STD    & MIN     & MAX    & SUM      \\ \midrule
Iteration \#1 & 0.8296 & 1.1596 & 1.0535 & -1.9828 & 1.9901 & 211.245  \\
Iteration \#2 & 0.832  & 1.1627 & 1.0746 & -1.9617 & 1.9946 & 212.991  \\
Iteration\#3  & 0.842  & 1.1495 & 1.0611 & -1.898  & 1.9998 & 213.5419 \\ \bottomrule
\end{tabular}
\caption{Various Statistical Parameters for weights obtained after regression for three different iterations. Each Iteration was run for 7560 epochs for improvement of Mean Square Error. Image used is 256x256 Baboon Image, noise levels as 0.09 for Iteration \#1 , 0.10 for Iteration\#2 and 0.11 for Iteration\#3 with $\eta$ = 10\%}
\label{my-label}
\end{table}
\begin{figure*}[t]
  \includegraphics[width=\textwidth,height=8cm]{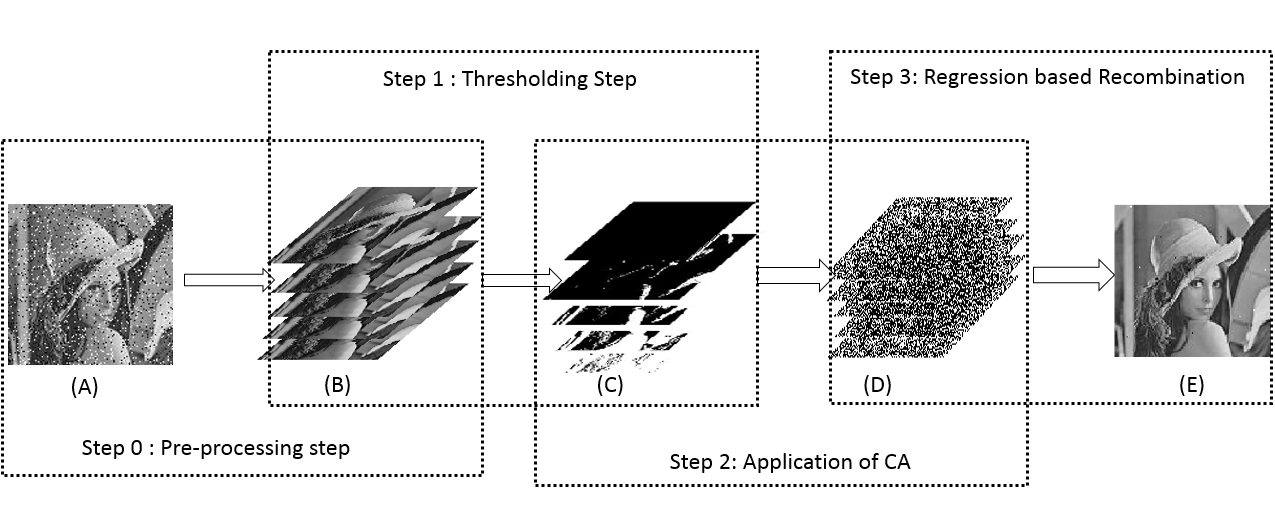}
  \caption{(A) is the original input image to the algorithm. Since for demonstration, we have chosen denoising Salt and Pepper noise, the illustration is a lenna image with 0.1 Salt and Pepper Noise. (B) is a 256-stack of the same image for thresholding in Step 1. (C) is the thresholded stack from (B). This can in general be a $\vert K \vert$-stack. (D) is stack after application of the Cellular Automaton on each of the binary images individually. The 3-step algorithm is essentially the generalization of this step to Gray-scale Images. (E) is the final output image after weighed recombination of individual images from (D) stack based on optimizing output PSNR.}
\end{figure*}
As can be seen from Fig.1 for $\eta$ as small a 1\% for a large enough image, the PSNR value does not depreciate much from what is achieved by using all uncorrupted pixels. This provides a speedup of two-orders for magnitude making the algorithm good enough for real-time implementation given a parallel architecture. But for smaller images (128x128 in size) for $\eta$ $\leq$ 10\% the number of samples become too less to have the same information as required and the performance suffers a lot. Thus, we choose $\eta = 10\%$ for other simulations also. \\
We have tested the algorithm for SPN noise varying from p = 0.01 to 0.2. At higher noise levels, the amount of uncorrupted pixels that can be reliably observed and used for regression, decreases. We have used Peak Signal to Noise Ratio (PSNR) as a metric to compare algorithm's performance.Given a noise-free m×n monochrome image I and its noisy approximation K, MSE is defined as \cite{wiki}
\begin{equation}
MSE = \frac{\sum\limits_{i=0}^{m-1}\sum\limits_{j=0}^{n-1}[I(i,j) - K(i,j)]^2}{mn}
\end{equation}
Consequently, Peak Signal to Noise Ratio (PSNR) is calculated as:
\begin{equation}
PSNR = 10\log_{10}\Big(\frac{255^2}{MSE}\Big)
\end{equation}
\\

Table 1 lists the various PSNR values achieved in the mid noise levels. Clearly, the SPN denoising algorithm performs better than the standard median filter consistently over various realizations and levels of noise. We have used thresholding at all possible values for testing and evaluation of the SPN denoising algorithm. Thus, setting the W vector identically to 1 is equivalent to circumventing the regression step and directly adding the binary images up in STEP 3. We conclude that as the noise present in the image increases more and more performance gain can be achieved by adjusting weights instead of empirically adding the binary images up. This is justified, since more and more distortion occurs from the uniform distribution of weights as the noise levels increase.  \\

Additionally, a clever trick that can be readily used for performance boost with absolutely minimal additional time and space requirements is to pre-calculate the weights for recombination in case of denoising SPN noise. If the noise induced in the image is expected to lie in the range of 0.05 - 0.20, then these weights can be reliably pre-calculated at the time of either generating the image or before sending it over a noisy channel. As can be seen from Table II, the actual distribution of weight remains practically the same and is not very much dependent on small changes in the noise level and on the actual realization of the noise with same noise level in the image. This allows the pre-processing of image at the time the image is generated and later  recombination within minimal time overheads. \\
Also, it is observed that the minimizing objective as defined by Eq.(5) is sufficiently nice and flat to allow for use of local minima in lieu of global minima finding algorithms. Specifically, instead of using methods that search for global minima over all possible space such as Simulated annealing, we can reasonably limit the search space to an appropriate value (depending on the number of thresholds used) and use local minima finding approaches on this smaller space for identical performance results and weight distributions.

\section{CONCLUSIONS}
We conclude by summarizing the three invariances discovered by our work and their possible exploitations which allow us to use regression based weighed recombination of binary images as a method for extending CA to grayscale images:
\begin{itemize}
\item Invariance to subsampling: This allows us to randomly uniformly sample a $\eta$\% of uncorrupted pixels for use in regression for finding the weights yielding a corresponding speedup in the process. $\eta$ can be taken to be as low as 1\% for large images (512x512 or larger) without any practical deterioration of performance.
\item Invariance to actual noise realization : For narrow ranges of noise level between 0.05 to 0.2, the actual distribution of weights can be pre-calculated and transmitted alongwith the image. This facilitates practically zero overhead in time for the end user using the algorithm along-with allowing use of CA for  gray-scale image processing in realtime. 
\item Equivalence of Local and Global Minima: This allows us to significantly cut down on the search space of possible weights. The algorithms for finding the local minima can be used instead of global minimal search algorithms such as  Simulated Annealing, further reducing the time for the one-time per image generation of weights.
\end{itemize}
Further optimization can be achieved by intializing the weights to an even better start or by developing a clever method for determining the thresholding weights and thereby, reducing the number of binary images that need to be recombined. Moreover, the use of inherent parallel structures of CA can be exploited to reduce the overall time overhead for application of CA to both binary and gray-scale images.

\addtolength{\textheight}{-12cm}   




\section{ACKNOWLEDGMENT}

The author would like to express his gratitude towards SURGE (Student Undergraduate Research Grant for
Excellence) programme offered by IIT Kanpur for the grant and the opportunity to pursue the project. It would not have been possible without the kind support and help of many
individuals and the staff of the dept. The author would like to extend his sincere thanks to all of them. 


\end{document}